\AddToHook{package/hyperref/before}{\RequirePackage{float}}
\documentclass[a4paper]{cas-dc}

\usepackage[authoryear,sort]{natbib}

\usepackage{amsmath,amssymb,amsfonts}
\makeatletter
\@fleqnfalse
\makeatother
\usepackage{graphicx}
\usepackage{textcomp}
\usepackage{algorithm}
\usepackage{algpseudocode}
\usepackage{multirow}
\usepackage{booktabs}
\usepackage{tabularx}
\usepackage{placeins}

\DeclareRobustCommand{\meanstd}[2]{#1\,$\pm$\,#2}
\DeclareRobustCommand{\tablestd}[2]{#1\scalebox{0.82}{{\color[gray]{0.45}(#2)}}}
\newcommand{\tablehead}[1]{{\fontsize{6.5}{7.5}\selectfont #1}}
\let\printorcid\relax
\providecommand{\IEEEPARstart}[2]{#1#2}
\let\cite\citep

\def\tsc#1{\csdef{#1}{\textsc{\lowercase{#1}}\xspace}}
\tsc{WGM}
\tsc{QE}

\ExplSyntaxOn
\cs_set:Npn \__first_footerline:
{
  \group_begin:
  \small\sffamily
  \ifnum\theblind>0\relax\else\__short_authors: :~\fi
  {\rmfamily\itshape Preprint}
  \group_end:
}
\ExplSyntaxOff

\begin{document}
\let\WriteBookmarks\relax
\def\floatpagepagefraction{1}
\def\textpagefraction{.001}

\shorttitle{ToRe}    

\shortauthors{Nanxi Yu et al.}  

\title [mode = title]{Learning When to Recur: Token-Adaptive Recursion for Imbalanced Ophthalmic Domain Incremental Learning}  

\author[2]{Nanxi Yu}

\author[1]{Kang Li}
\cormark[1]
\ead{kangli@uestc.edu.cn}

\author[2]{Ye Du}

\author[3]{Xiaowei Hu}

\author[4]{Weihua Yang}

\author[2]{Shujun Wang}
\cormark[1]
\ead{shu-jun.wang@polyu.edu.hk}

\affiliation[1]{organization={School of Mechanical and Electrical Engineering, University of Electronic Science and Technology of China},
            state={Sichuan},
            country={China}}
\affiliation[2]{organization={Department of Biomedical Engineering and Sports Technology, The Hong Kong Polytechnic University},
            country={Hong Kong SAR, China}}
\affiliation[3]{organization={School of Future Technology, South China University of Technology},
            state={Guangdong},
            country={China}}
\affiliation[4]{organization={Shenzhen Eye Hospital, Shenzhen Eye Medical Center, Southern Medical University},
            state={Guangdong},
            country={China}}

\cortext[1]{Corresponding authors.}

\begin{abstract}
Domain incremental learning is essential for adapting ophthalmic deep learning models to sequential clinical domains while preserving diagnostic expertise. 
Existing domain incremental learning methods predominantly address the domain shift induced by style variations. 
However, they often overlook the severe class imbalance inherent in real-world clinical scenarios, such as clinical referral systems.
Institutions in these systems encounter drastic fluctuations in class priors, resulting in label distribution shift, a critical form of domain shift that triggers severe catastrophic forgetting.
To address these challenges, we propose ToRe, a rehearsal-free and parameter-efficient framework that leverages frozen ophthalmic foundation models for robust incremental adaptation. ToRe employs a parameter isolation strategy to decouple domain-specific optimization paths, thereby helping mitigate catastrophic forgetting driven by both label distribution shift and style variations.
Simultaneously, it introduces token-adaptive recursion that adaptively allocates additional computational depth across tokens, allowing simple tokens to exit the recursion loop early while subjecting complex tokens, such as those associated with lesions, to deeper recursive processing.
This mechanism enhances the feature representations for minority classes, thereby supporting generalization throughout the domain incremental learning process.
Extensive evaluations on nine heterogeneous datasets demonstrate that ToRe consistently outperforms state-of-the-art methods in overall performance across the three benchmarks, while maintaining near-zero forgetting. Together, these results support the applicability of ToRe to dynamic and imbalanced clinical environments.
The code is available at \url{https://github.com/Nancyolo/ToRe}.
\end{abstract}


\maketitle

\section{Introduction}
\label{sec:introduction}

\IEEEPARstart{O}{phthalmic} diseases, such as diabetic retinopathy (DR) and age-related macular degeneration (AMD), constitute a primary cause of preventable blindness globally~\cite{steinmetz2021causes}. 
While deep learning has shown remarkable potential in diagnosing these diseases~\cite{gulshan2016development,burlina2017automated}, its clinical utility is often limited by the nonstationary nature of real-world environments~\cite{lee2020clinical,parisi2019continual}.
In practice, ophthalmic data streams originate sequentially from various medical institutions with heterogeneous imaging devices, acquisition protocols, and patient demographics, collectively characterizing a pronounced \emph{\textbf{domain shift}}~\cite{jiang2017automatic,bhati2023discriminative}.
Consequently, an ideal diagnostic model requires the capability to continually adapt to these new domains (e.g., centers), without losing previously acquired expertise.
To this end, \emph{domain incremental learning (DIL)} has emerged as a critical paradigm for cumulative knowledge acquisition across diverse clinical settings.~\cite{van2019three}.
\par
\suppressfloats[t]
\begin{figure}[pos=!t]
\centering
\includegraphics[width=\columnwidth]{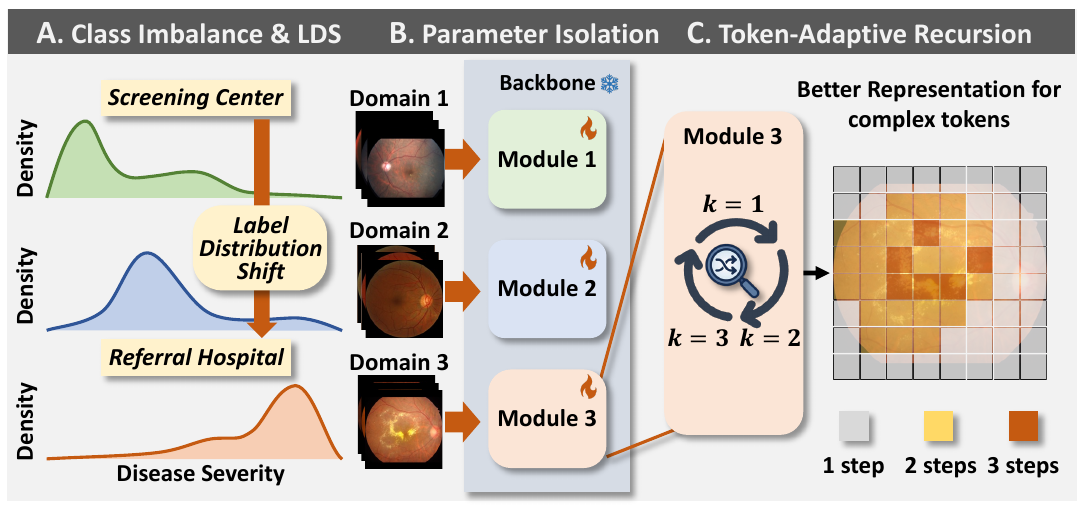}
\caption{\textbf{Conceptual overview of ToRe for imbalanced ophthalmic DIL.}
(A) \textbf{Challenges}: Fluctuating class priors across screening and referral institutions create severe imbalance and \textit{label distribution shift} (LDS), leading to severe catastrophic forgetting.
(B) \textbf{Parameter Isolation}: Domain-specific paths are optimized independently to reduce interference driven by LDS and style variations.
(C) \textbf{Token-Adaptive Recursion}: Breaking the uniform depth constraint of standard Vision Transformers, this mechanism adaptively increases computational depth for complex tokens, such as those associated with small lesions, to support representation learning for minority classes.
}
\label{fig:teaser_concept}
\end{figure}

The primary challenge in DIL is \textbf{\textit{catastrophic forgetting}}, where adapting to new domains degrades performance on previous ones due to the aforementioned domain shift~\cite{li2017learning,li2022domain}.
To mitigate this, existing DIL strategies primarily rely on regularization-based~\cite{kirkpatrick2017overcoming,li2017learning} or replay-based~\cite{lopez2017gradient,chaudhry2019continual,chrysakis2020online,kim2020imbalanced} approaches, which have been actively explored for medical imaging~\cite{wu2024continual,li2022domain, li2024dual, perkonigg2021dynamic, srivastava2021continual}. 
Despite their effectiveness, these approaches are often inapplicable to practical clinical deployment, limited by two critical bottlenecks.
First, data privacy regulations can restrict the construction of replay buffers~\cite{verma2023privacy}. 
Second, the computational burden of full fine-tuning or computing complex regularization terms is often prohibitive for local clinical deployment~\cite{li2017learning}.
Consequently, these constraints necessitate a transition to a rehearsal-free, parameter-efficient fine-tuning (PEFT) paradigm~\cite{hu2022lora,jia2022visual,lester2021power,houlsby2019parameter}. 
Specifically, PEFT enables model adaptation by optimizing only a small subset of parameters while keeping the backbone frozen.

Current PEFT-based DIL methods primarily focus on addressing the domain shift induced by style variations, such as image contrast variations among fundus camera vendors~\cite{wang2022dualprompt,wang2022s,smith2023coda,wang2022learning,wang2025hide,zhu2024memory,10981824}.
However, these methods exhibit critical vulnerabilities in ophthalmic scenarios due to two primary gaps:
(i) a lack of specialized foundational representations tailored for retinal features to initialize model generalization capability~\cite {zhou2023foundation, chia2024foundation, shi2024eyefound},
and (ii) more importantly, the neglect of an essential aspect: the severe \textbf{\emph{class imbalance}} and the resulting \textbf{\emph{label distribution shift (LDS)}}, a critical form of domain shift that is frequently observed in clinical referral systems.
As illustrated in Fig.~\ref{fig:teaser_concept} A, distributions are naturally skewed across institutions: screening centers are normally dominated by healthy cases, whereas specialized referral hospitals receive a high concentration of minority classes~\cite{aptos2019,decenciere2014feedback,porwal2018indian}. 
This causes class priors to fluctuate drastically, creating the LDS phenomenon. 

Overcoming these challenges remains difficult given the structural conventions of current PEFT-based DIL models.
These models typically adopt standard vision transformers (ViTs) as their backbones, which process all image tokens with uniform depth, regardless of their semantic importance~\cite{rao2021dynamicvit, yin2022vit}. 
In imbalanced ophthalmic data, this uniformity causes the model to favor abundant healthy regions, whereas sparse lesions are left with insufficient processing.
When further tackling these varying class imbalance situations in sequentially arriving domains, the lack of specific precautions for LDS causes the model to overfit the current majority class.
Consequently, the fragile representations of previously learned minority classes are easily overwritten, triggering catastrophic forgetting.

To address catastrophic forgetting while improving diagnostic performance under class imbalance and the resulting LDS, we propose \textbf{ToRe}, a PEFT framework that leverages frozen ophthalmic foundation models (FMs) to tackle both the LDS and style variations across domains.
Our approach adopts a parameter isolation paradigm, assigning independent optimization paths for domain-specific modules to decouple the learning process of various domains (Fig.~\ref{fig:teaser_concept} B).
Within each domain, we introduce \emph{\textbf{token-adaptive recursion}} to address the class imbalance challenge by learning which tokens should continue through the recursion loop.
By breaking the uniform depth constraint of standard ViT, this mechanism adaptively allocates computational depth: simple tokens exit the recursion loop early, whereas complex tokens, such as those associated with small lesions, undergo recursive processing (Fig.~\ref{fig:teaser_concept} C).
This improves discriminative diagnostic representations under class imbalance.
At inference, we employ feature-based domain identification to select the matching domain-specific module for prediction.
Together, parameter isolation and domain identification help preserve and retrieve the discriminative knowledge acquired via token-adaptive recursion under LDS and style variations across domains, thereby mitigating catastrophic forgetting and improving diagnostic performance under class imbalance.
In summary, our main contributions are summarized as follows:
\begin{itemize}
    \item We propose \textbf{ToRe}, a novel rehearsal-free DIL framework that combines token-adaptive recursion with parameter isolation to address class imbalance and LDS.

    \item We employ a parameter isolation strategy to reduce optimization interference across domains and a domain identification mechanism to select the module that best matches the input feature representation at inference. Together, they preserve and retrieve domain-specific knowledge under LDS and style variations.

    \item We introduce \textbf{token-adaptive recursion} that adaptively increases the \textbf{computational depth} for complex tokens. By breaking the uniform depth constraint of standard ViTs, this mechanism addresses class imbalance by improving the representation of minority classes.

    \item We conduct extensive evaluations on three ophthalmic benchmarks. ToRe consistently outperforms state-of-the-art methods in overall performance across these benchmarks. Notably, it demonstrates strong diagnostic performance under severe class imbalance while achieving near-zero forgetting.

\end{itemize}

\section{Related Work}
\label{sec:relatedwork}
This section reviews literature across three relevant research streams. We first survey domain incremental learning in Sec.~\ref{Related:DIL}. Building on this, Sec.~\ref{Related:imbalance} examines the challenge of class imbalance in sequential streams. Finally, Sec.~\ref{Related:DNN} explores adaptive computation and token-dependent recursion.
\subsection{Domain Incremental Learning}
\label{Related:DIL}
\subsubsection{Traditional Methods for Medical DIL}
\begingroup\emergencystretch=1em
DIL addresses scenarios where the label space remains fixed, whereas the input distribution shifts sequentially~\cite{wu2024continual}.
Traditional approaches mitigate catastrophic forgetting through mechanisms such as regularization~\cite{kirkpatrick2017overcoming,aljundi2018memory}, replay~\cite{lopez2017gradient,chaudhry2019continual}, and distillation~\cite{li2017learning,rebuffi2017icarl}.
In the medical domain, extensive studies have adapted these paradigms across modalities~\cite{wu2024continual,li2022domain, li2024dual, gonzalez2023lifelong, perkonigg2021dynamic, zhang2023continual, derakhshani2022lifelonger, srivastava2021continual, bayasi2024biaspruner}.
Specifically, prominent frameworks rely heavily on {experience replay} or maintaining {dynamic memory banks} to preserve historical knowledge.
Other approaches employ regularization or pruning techniques~\cite{bayasi2024biaspruner} to mitigate bias without storing raw data.
Although effective in research settings, these heavy frameworks typically require accessing historical patient data or substantial computation.
This dependency violates ophthalmic data privacy regulations and compromises the computational efficiency required for local clinical deployment.
\par\endgroup
\subsubsection{Parameter-Efficient Fine-Tuning for DIL}
\label{Related:PEFT}
To circumvent these barriers, PEFT has emerged as a dominant solution.
Rather than full fine-tuning, PEFT strategies inject lightweight modules, such as LoRA~\cite{hu2022lora}, Prompts~\cite{wang2022learning}, or Adapters~\cite{zhang2023adapter}, into a frozen backbone to facilitate adaptation.
Existing PEFT-based DIL approaches generally fall into two categories: parameter sharing and parameter isolation.
\textit{Parameter sharing} strategies (e.g., L2P~\cite{wang2022learning}, DualPrompt~\cite{wang2022dualprompt}, CODA-Prompt~\cite{smith2023coda}, HiDe-PET~\cite{wang2025hide}) maintain shared parameters like prompt pools to facilitate knowledge transfer.
However, under LDS, optimizing shared parameters can modify representations used by earlier domains and thereby create cross-domain interference.
Conversely, {parameter isolation} strategies (e.g., S-Prompts~\cite{wang2022s}) instantiate independent modules for each domain to decouple optimization.
S-Prompts learns prompts independently across domains and uses K-NN to identify the domain of each test sample during inference.
Although recent medical adaptations have explored this paradigm to handle domain shift induced by style variations (e.g., histopathology stain differences~\cite{zhu2024memory}), they overlook the critical impact of class imbalance in retinal streams. In such imbalanced scenarios, standard isolated parameters often fail to robustly capture sparse features, creating a performance bottleneck.

\subsection{Class Imbalance in Sequential Settings}
\label{Related:imbalance}
Addressing class imbalance in sequential streams is traditionally achieved by manipulating the replay buffer to enforce balanced storage~\cite{chrysakis2020online, kim2020imbalanced}.
Other approaches optimize buffer update policies using KL-divergence losses~\cite{nikoloutsopoulos2022online} or employ two-stage training with balanced sampling~\cite{liu2022long}.
Essentially, these methods rely on physically re-balancing the training distribution via memory.
However, in our rehearsal-free setting mandated by privacy, such buffer-based strategies are inapplicable.
Alternative strategies like Logit Adjustment~\cite{menon2020long}, although buffer-free, are designed for static long-tailed recognition and fail to address the catastrophic forgetting inherent to DIL.
Recent attempts like DCE~\cite{li2025addressing} try to mitigate this by calibrating classifiers, but relying on a static feature space limits their ability to robustly process minority classes across shifting domains.
This necessitates a mechanism to dynamically enhance the quality of feature representations directly within the feature extractor.
\begin{figure*}[pos=t!]
    \centering
    \includegraphics[width=\linewidth]{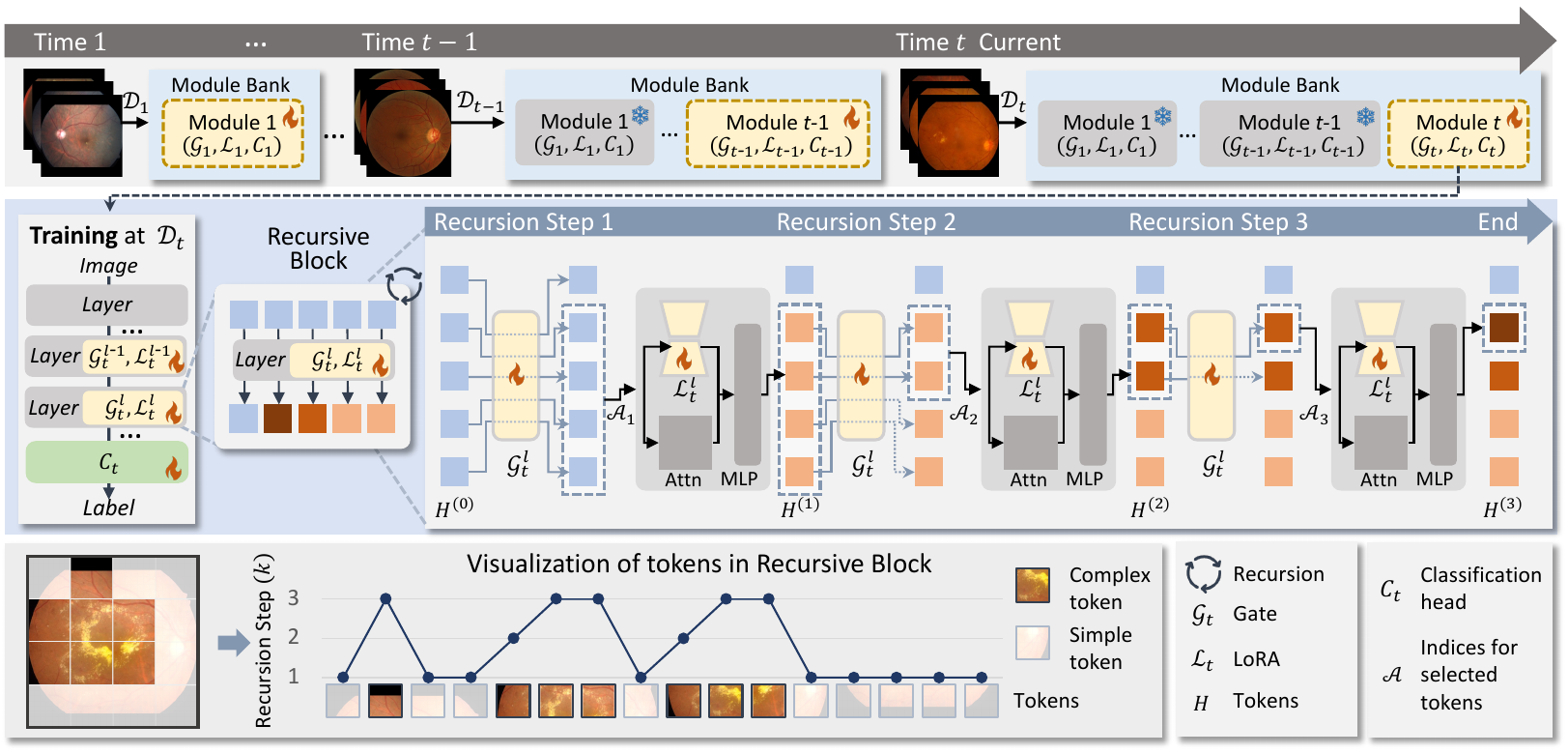}
    \caption{\textbf{Training Process of ToRe.}
    \textbf{(Top) Sequential Domain Stream:} The model learns from a sequence of domains ($\mathcal{D}_1 \dots \mathcal{D}_t$) via \textbf{parameter isolation}. For each new domain $\mathcal{D}_t$, a specific set of modules (Gating Module $\mathcal{G}_t$, LoRA $\mathcal{L}_t$, and Classification head $\mathcal{C}_t$) is initialized and trained, whereas previous modules remain frozen. After training, the domain-specific module and its domain key are registered in the module bank.
    \textbf{(Middle) Training at $\mathcal{D}_t$:} The left panel illustrates the model architecture for domain $\mathcal{D}_t$. The right panel unrolls a single \textbf{recursive block} with recursion steps 1--3 ($K_{\max}=3$).
    \textbf{(Bottom) Visualization of token dynamics in a single recursive block:} At each recursion step, the gating module selects tokens for recursive processing. Selected tokens receive recursive feature updates, while unselected tokens retain their current states.}
    \label{fig:DIL}
\end{figure*}
\begin{figure}[pos=t!]
\centering
\includegraphics[width=\linewidth]{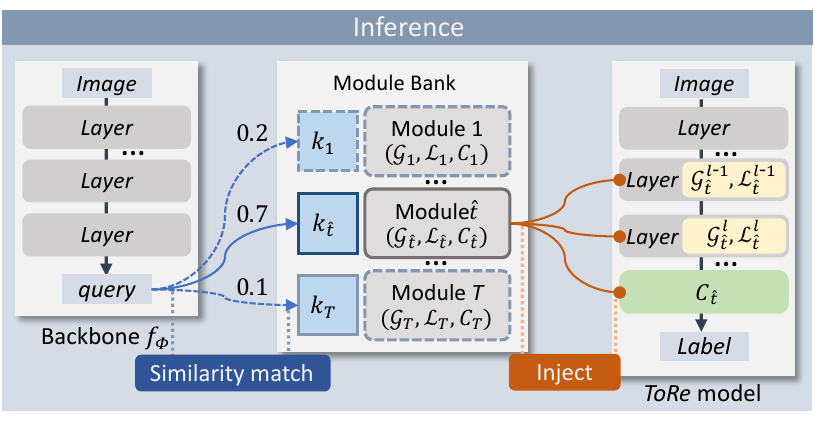}
\caption{\textbf{Inference Flow with Domain Identification.} At inference, a query vector is extracted by the frozen backbone $f_{\Phi}$ and matched against pre-computed domain keys ($k_1 \dots k_T$) in the module bank via cosine similarity. The module ($\Theta_{\hat{t}} = \{\mathcal{G}_{\hat{t}}, \mathcal{L}_{\hat{t}}, \mathcal{C}_{\hat{t}}\}$) corresponding to the highest similarity score is retrieved and injected to perform the final diagnostic prediction.}
\label{fig:inference_flow}
\end{figure}
\subsection{Adaptive Computation}
\label{Related:DNN}
Adaptive computation aims to allocate resources conditioned on input complexity.
Early works like the Mixture-of-Experts~\cite{shazeer2017outrageously} introduced conditional routing to activate specific expert subsets.
Subsequent research in vision transformers optimized efficiency through dynamic token reduction.
Methods like DynamicViT~\cite{rao2021dynamicvit} and A-ViT~\cite{yin2022vit} learn to progressively prune uninformative tokens, whereas token merging~\cite{bolya2022token} aggregates redundant features.
Recently, the Mixture-of-Recursions~\cite{bae2025mixture} extended this idea to token-dependent recursion, allowing complex tokens to undergo deeper recursive processing.
Our work draws inspiration from this concept but adapts it to address the class imbalance challenge.

\section{Methodology}
\label{sec:method}

In this section, we propose \textbf{ToRe}, a rehearsal-free framework tailored for ophthalmic domain incremental learning.
We first formalize the ophthalmic DIL problem in Sec.~\ref{sec:problem}, identifying the specific challenges of class imbalance and distribution shifts. Building on this, Sec.~\ref{sec:framework} details our framework architecture, and Sec.~\ref{sec:reclora} introduces the core token-adaptive recursion, which provides adaptive computational depth for complex token features.

\subsection{Problem Formulation and Analysis}
\label{sec:problem}
\subsubsection{DIL Setup}
We consider the DIL setting where data streams arrive from $T$ distinct domains, denoted as $\mathcal{S} = \{\mathcal{D}_1, \dots, \mathcal{D}_T\}$. At any time step $t \in \{1, \dots, T\}$, the model only has access to the current domain $\mathcal{D}_t = \{(x_i, y_i)\}_{i=1}^{N_t}$, which consists of retinal images $x \in \mathcal{X}$ and diagnostic labels $y \in \mathcal{Y}$, represented as a class index for the single-label benchmarks or a multi-hot vector for the multi-label benchmark.
Consistent with the DIL paradigm, the label space $\mathcal{Y}$ remains fixed across domains, whereas the underlying data statistics vary due to the nonstationary clinical environment. 
Ideally, DIL aims to minimize the cumulative risk across all domains encountered up to the current time $t$, with $P_k(X, Y)$ denoting the joint distribution of the $k$-th domain:
\begin{equation}
\label{eq:joint_optimization}
\theta^* = \mathop{\arg\min}_{\theta} \sum_{k=1}^{t} \mathbb{E}_{(x,y) \sim P_k(X, Y)} \big[\mathcal{L}(f_\theta(x), y)\big].
\end{equation}
In practice, the sequential nature of domain arrival and the lack of a rehearsal buffer prevent the direct joint optimization of Eq.~\eqref{eq:joint_optimization}, necessitating domain incremental learning strategies.
\subsubsection{Analysis of the Compound Challenge}
Under these sequential-access and no-rehearsal constraints, optimization at step $t$ is restricted to the current-domain objective:
\begin{equation}
\label{eq:decomposed_optimization}
\theta_t = \mathop{\arg\min}_{\theta} \mathbb{E}_{y \sim P_t(Y)} \Big[ \mathbb{E}_{x \sim P_t(X|y)} \big[\mathcal{L}(f_\theta(x), y)\big] \Big].
\end{equation}
To understand why current PEFT methods struggle in ophthalmic DIL, we decompose the joint distribution of the current domain into two components: $P_t(X, Y) = P_t(X|Y) \cdot P_t(Y)$. This decomposition highlights the dual nature of domain shifts.
The first component, $P_t(X|Y)$, represents the domain shift on the input side (i.e., style variations), caused by variations in imaging devices.
More critically, the second component, the label distribution $P_t(Y)$, introduces class imbalance and label distribution shift, which represents the domain shift on the label side.
Here, the outer expectation $\mathbb{E}_{y \sim P_t(Y)}$ implies that the loss is explicitly weighted by the label distribution of the current domain.
Together, changes in the label distribution $P_t(Y)$ and conditional input distribution $P_t(X|Y)$ create two coupled optimization challenges.
First, the skew in the label distribution (class imbalance) leads to what we term representation deficiency: as training is dominated by majority samples, minority-class representations can remain insufficiently learned.
Second, updating shared parameters across domains under these distribution changes can alter parameters used by earlier domains and create cross-domain interference.
These coupled challenges motivate us to adopt a parameter-isolation strategy to separate domain-specific optimization pathways, while the token-adaptive recursion refines minority-class representations.

\subsection{Overall Framework}
\label{sec:framework}
To address the limitations of ophthalmic DIL identified in Sec.~\ref{sec:problem}, we propose the framework, ToRe.
As illustrated in Fig.~\ref{fig:DIL} and Fig.~\ref{fig:inference_flow}, the overall architecture comprises three architectural elements:
1) A frozen backbone derived from a foundation model to extract general features;
2) a module bank that stores isolated, domain-specific modules to reduce cross-domain interference;
3) recursive blocks that inject these domain-specific modules into the backbone to enable token-adaptive recursion for feature updates.
In the following, we detail the architectural composition (Sec.~\ref{sec:framework_arch}), training with parameter isolation (Sec.~\ref{sec:framework_opt}), and the inference mechanism (Sec.~\ref{sec:framework_inf}).

\subsubsection{Framework Architecture}
\label{sec:framework_arch}
We construct our framework upon a backbone $f_{\Phi}$, initialized from an ophthalmic foundation model (e.g., RETFound~\cite{zhou2023foundation}).
Adhering to the PEFT-based DIL paradigm, we freeze all backbone parameters $\Phi$ to preserve their retinal feature representations.
Simultaneously, to enable adaptation to sequential domains, we introduce lightweight trainable modules to capture domain-specific knowledge.
Structurally, these modules are organized within a \textit{module bank} and injected into the backbone via \textit{recursive blocks} to perform the diagnostic classification.

\textit{\textbf{module bank ($\mathbb{M}$):}}
To reduce cross-domain interference, we establish a \textit{module bank} $\mathbb{M} = \{\Theta_1, \dots, \Theta_T\}$, acting as a repository of domain-specific modules.
As shown in Fig.~\ref{fig:DIL} (Top), for each incoming domain $\mathcal{D}_t$, we instantiate a new set of trainable parameters $\Theta_t = \{\mathcal{G}_t, \mathcal{L}_t, \mathcal{C}_t\}$, which contain gating modules $\mathcal{G}_t$, LoRA parameters $\mathcal{L}_t$, and a classifier head $\mathcal{C}_t$.
By assigning a separate parameter set to each domain, we separate the optimization pathways. This design addresses the challenge discussed in Sec.~\ref{sec:problem}: under changes in class priors $P_t(Y)$ and class-conditional input distributions $P_t(X|Y)$, current-domain optimization updates only the current module $\Theta_t$, leaving previously learned modules unchanged. Consequently, optimization for each new domain can focus on improving the diagnostic performance of its domain-specific module under class imbalance.

\textit{\textbf{Recursive Block Injection:}}
To address representation deficiency, we inject the current module $\Theta_t$ during training or the identified module $\Theta_{\hat{t}}$ during inference into the specific backbone layers, enabling the model to recursively update the representations of tokens selected by the domain-specific module.
This injection reconfigures backbone layers into \textit{recursive blocks}.
As detailed in Fig.~\ref{fig:DIL} (Middle), inside a recursive block, the frozen layer provides base feature extraction, whereas the injected domain-specific module introduces a learnable recursion loop (Sec.~\ref{sec:reclora}).
This mechanism grants the model adaptive computational depth, improving class-discriminative diagnostic representations while the backbone parameters remain fixed.
\subsubsection{Training with Parameter Isolation}
\label{sec:framework_opt}
During current-domain training, the instantiated module $\Theta_t$ is injected into the designated recursive blocks and optimized, while the backbone $\Phi$ and previous modules remain fixed.
For the current domain $\mathcal{D}_t$, we optimize only the learnable parameters $\Theta_t$ under the local decomposed distribution:
\begin{equation}
\label{eq:local_optimization}
\Theta_t^* = \mathop{\arg\min}_{\Theta_t} \mathbb{E}_{y \sim P_t(Y)} \Big[ \mathbb{E}_{x \sim P_t(X|y)} \big[\mathcal{L}(f_{\Phi}(x; \Theta_t), y)\big] \Big].
\end{equation}

\subsubsection{Inference with Domain Identification}
\label{sec:framework_inf}
During inference, the domain identity of a test sample $x_{\text{new}}$ is unknown.
We employ a domain-identification mechanism that leverages the feature space of the frozen backbone (Fig.~\ref{fig:inference_flow}).

\textit{\textbf{Domain Key Construction:}} After training on domain $\mathcal{D}_t$, we compute a \textit{domain key} $k_t$, defined as the centroid of the domain's training samples in the frozen feature space:
\begin{equation}
k_t = \frac{1}{|\mathcal{D}_t|} \sum_{x \in \mathcal{D}_t} f_{\Phi}(x).
\end{equation}
We then register the association $(k_t,\Theta_t)$ in the module bank.

\textit{\textbf{Similarity Matching:}} For a test sample, we extract its query vector $q = f_{\Phi}(x_{\text{new}})$ and identify the most relevant domain index $\hat{t}$ via cosine similarity:
\begin{equation}
\hat{t} = \mathop{\arg\max}_{j \in \{1, \dots, T\}} \frac{q \cdot k_j}{\|q\| \|k_j\|}.
\end{equation}
As shown in Fig.~\ref{fig:inference_flow}, we then retrieve the module $\Theta_{\hat{t}}$ from the bank and inject it into the backbone to perform the final diagnostic prediction.
\begin{algorithm}[t]
\color{black}
\small
\caption{Recursive Forward Pass in a Recursive Block}
\label{alg:reclora}
\begin{algorithmic}[1]
\State \textbf{Input:} Token states $H^{(0)} \in \mathbb{R}^{N \times d}$, Max depth $K_{\text{max}}$
\State \textbf{Params:} Frozen $\Phi$, Trainable $\Theta_{t} = \{\mathcal{G}_{t}, \mathcal{L}_{t}\}$, Embeddings $\{E_{\text{iter}}^k\}$
\State Init candidate indices $\mathcal{A}_0 \leftarrow \{1, \dots, N\}$
\For{$k = 1$ \textbf{to} $K_{\text{max}}$}
    \State \textbf{1. Embed:} $H_{\text{in}} \leftarrow H^{(k-1)} + E_{\text{iter}}^k$
    \State \textbf{2. Gate:} Compute mask $M^{(k)}$ via Eq.~\eqref{eq:gating} using $H_{\text{in}}$
    \State \quad Update selected set $\mathcal{A}_k \leftarrow \{ i \in \mathcal{A}_{k-1} \mid M_i^{(k)} = 1 \}$
    \State \quad \textbf{if} $\mathcal{A}_k = \emptyset$ \textbf{break}
    \State \textbf{3. Update:} Compute $H^{(k)}_{\mathcal{A}_k}$ via Eq.~\eqref{eq:update_rule}
    \State \quad $H^{(k)}_{\notin \mathcal{A}_k} \leftarrow H^{(k-1)}_{\notin \mathcal{A}_k}$ (Skip unselected)
\EndFor
\State \textbf{Return} $H^{(k)}$
\end{algorithmic}
\end{algorithm}
\subsection{Token-Adaptive Recursion}
\label{sec:reclora}

To address the representation deficiency of minority classes, we focus on a critical question: \textit{how to effectively improve representation for minority classes under severe class imbalance?}
We observe that standard ViTs process all image patches with uniform computational depth. In ophthalmic diagnosis, this uniformity contributes to representation deficiency. The vast majority of tokens represent healthy tissue (e.g., optic disc), which can contain relatively simple and abundant visual patterns. Conversely, lesion-associated tokens (e.g., those containing exudates) can carry complex diagnostic features, yet they receive the same computational depth as background tokens.
Consequently, the model is dominated by the easy, majority patterns, leaving critical signatures of minority classes under-represented.

To break this uniformity constraint, we propose \textit{token-adaptive recursion}. 
Its core intuition is to allocate additional computation based on feature complexity: complex tokens receive recursive updates, while simple tokens retain their previous states.
To build on the pretrained representations of the frozen backbone, the recursion loop in each recursive block starts from the token representations initially produced by its LoRA-augmented layer.
For example, as visualized in Fig.~\ref{fig:DIL} (Bottom), a lesion-associated token selected for recursion can undergo deeper recursive processing over recursion steps 1--3, whereas unselected tokens retain their current states. This strategy allocates additional computation to improve discriminative representations under class imbalance.

\subsubsection{Structural Composition}
To implement this logic within a foundation model, we reconfigure specific backbone layers into \textit{recursive blocks}. Each block integrates two functional components into the frozen architecture.
First, the \textit{gating module ($\mathcal{G}_t$)} acts as the ``selector.'' It is a lightweight linear projection that learns a token-selection policy through the classification task loss.
Second, the \textit{LoRA-augmented layer} acts as the ``refiner.'' It integrates trainable LoRA parameters $\mathcal{L}_t$ into the frozen attention mechanism, allowing the model to perform domain-specific feature updates while retaining the robustness of the foundation model.

\subsubsection{The Recursion Loop}
As illustrated in Fig.~\ref{fig:DIL} (Middle), the recursion loop executes a \textit{Select-and-Update} cycle for a maximum of $K_{\text{max}}$ steps. Let $H \in \mathbb{R}^{(N+1)\times d}$ denote the current token states, with index 0 denoting the class token (CLS) and indices $1,\dots,N$ denoting image patch tokens. We initialize the candidate set as $\mathcal{A}_0=\{1,\dots,N\}$.

\textit{\textbf{Token-Adaptive Gating:}}
At recursion step $k$, the gate determines which tokens continue through the recursion loop based on their current states $H$.
Since the gating module shares parameters across recursion steps within the same block and domain, it does not explicitly receive the current step index.
Therefore, to enable the gate to perceive the current recursion step, we inject a learnable \textit{iteration embedding} $E_{\text{iter}}^k$ into the input.
The gate then computes a probability map. We employ the \textit{Straight-Through Estimator (STE)}~\cite{yin2019understanding} to train the gate from the current-domain classification loss through discrete token selection:
\begin{equation}
\label{eq:gating}
M^{(k)} = \text{STE}\left(\sigma\left(\mathcal{G}_t(H + E_{\text{iter}}^k)\right)\right),
\end{equation}
where $\sigma(\cdot)$ is the sigmoid function. 
The selected set is $\mathcal{A}_k=\{i\in\mathcal{A}_{k-1}\mid M_i^{(k)}=1\}$. A token can be selected only if it belonged to the preceding candidate set, so $\mathcal{A}_k\subseteq\mathcal{A}_{k-1}$ and the set size is non-increasing. If no tokens are selected, the current recursion loop ends.

\textbf{\textit{Recursive Feature Updating:}}
Once the selected set $\mathcal{A}_k$ is identified, we apply a selective update. 
The selected tokens and CLS are processed by the LoRA-augmented layer with the current iteration embedding. Denoting its input token states by $H$, the update rule is defined as:
\begin{equation}
\label{eq:update_rule}
H \leftarrow \text{Layer}\left(H; \Phi, \mathcal{L}_t\right).
\end{equation}
Here, $\text{Layer}(\cdot)$ denotes the full computation of the transformer layer. Through this recursive pass, the LoRA-aug\-mented layer progressively updates the features of selected tokens. 
Unselected tokens retain their current states and do not participate as keys or values in the current recursion step. This recursion loop continues until $K_{\text{max}}$ is reached or no tokens are selected. All token states then proceed to the next backbone block. The recursion loop is summarized in Algorithm~\ref{alg:reclora}.

\newcommand{\tablecaption}[1]{#1}

\begin{table}[pos=t!]
\centering
\caption{\tablecaption{Statistics of class and label distributions across domains. For OCT, IRF, SRF, and PED denote per-label positive B-scan counts. The labels are non-exclusive, and Total denotes all B-scans, including those without any of the three target biomarkers.}}
\label{tab:dataset_stats}
\centering
\begingroup
\setlength{\tabcolsep}{1.5pt}
\makebox[\columnwidth][c]{%
\begin{tabular*}{\columnwidth}{@{\extracolsep{\fill}}lccccccc@{}}
\toprule
{Bench.} & {Domain} & {C0} & {C1} & {C2} & {C3} & {C4} & {Total} \\
\midrule
\multirow{3}{*}{DR} 
& APTOS & 1805 & 370 & 999 & 193 & 295 & 3662 \\
& Messidor-2 & 1017 & 270 & 347 & 75 & 35 & 1744 \\
& IDRiD & 168 & 25 & 168 & 93 & 62 & 516 \\
\midrule
{Bench.} & {Domain} & \multicolumn{2}{c}{{Normal}} & \multicolumn{3}{c}{{AMD}} & {Total} \\
\midrule
\multirow{3}{*}{AMD} 
& ODIR & \multicolumn{2}{c}{2290} & \multicolumn{3}{c}{207} & 2497 \\
& ADAM & \multicolumn{2}{c}{622} & \multicolumn{3}{c}{178} & 800 \\
& HYAMD & \multicolumn{2}{c}{799} & \multicolumn{3}{c}{415} & 1214 \\
\midrule
{Bench.} & {Domain} & \multicolumn{2}{c}{{IRF}} & \multicolumn{2}{c}{{SRF}} & {PED} & {Total} \\
\midrule
\multirow{3}{*}{OCT}
& RETOUCH & \multicolumn{2}{c}{2128} & \multicolumn{2}{c}{1427} & 1438 & 6936 \\
& AMD-SD & \multicolumn{2}{c}{2084} & \multicolumn{2}{c}{1789} & 727 & 3049 \\
& APTOS-2021 & \multicolumn{2}{c}{2399} & \multicolumn{2}{c}{971} & 525 & 2850 \\
\bottomrule
\end{tabular*}%
}
\endgroup
\end{table}

\begin{table*}[pos=t]
\centering
\caption{\tablecaption{Performance comparison on the \textbf{DR Grading} benchmark. Per-domain accuracy and overall metrics are reported as mean (std) over three seeds. Best results are in \textbf{bold}, second best are \underline{underlined}.}}
\label{tab:main_results_dr}
\centering
\shortcites{kirkpatrick2017overcoming,li2017learning,wang2022learning,wang2022dualprompt,wang2022s,smith2023coda,wang2025hide,li2025addressing}
\setlength{\tabcolsep}{0.4pt}
\fontsize{7.5}{9}\selectfont
\let\meanstd\tablestd
\begin{tabular*}{\textwidth}{@{\extracolsep{\fill}}l ccc ccccc@{}}
\toprule
 & \multicolumn{3}{c}{Per-Domain Accuracy (\%)} & \multicolumn{5}{c}{Overall Metrics} \\
\cmidrule(lr){2-4} \cmidrule(lr){5-9}
\tablehead{{Method}} & \tablehead{{APTOS}} & \tablehead{{Messidor-2}} & \tablehead{{IDRiD}} & \tablehead{{Avg.ACC (\%) $\uparrow$}} & \tablehead{{Avg. F1 (\%) $\uparrow$}} & \tablehead{{Avg. BAcc (\%) $\uparrow$}} & \tablehead{{Avg. MCC $\uparrow$}} & \tablehead{{Forgetting (\%) $\downarrow$}} \\
\midrule
FT-Seq & \meanstd{60.79}{1.48} & \underline{\meanstd{66.22}{0.48}} & \meanstd{52.56}{2.42} & \meanstd{59.86}{0.30} & \meanstd{46.40}{1.45} & \meanstd{46.75}{1.31} & \meanstd{0.40}{0.01} & \meanstd{13.25}{1.07} \\
EWC \cite{kirkpatrick2017overcoming} & \meanstd{64.24}{5.02} & \meanstd{63.44}{1.81} & \meanstd{52.56}{7.22} & \meanstd{60.08}{4.19} & \meanstd{44.12}{0.44} & \meanstd{44.16}{0.41} & \meanstd{0.41}{0.04} & \meanstd{10.49}{3.41} \\
LwF \cite{li2017learning} & \meanstd{68.18}{7.27} & \meanstd{65.02}{1.33} & \meanstd{49.68}{3.89} & \meanstd{60.96}{3.74} & \meanstd{41.71}{3.83} & \meanstd{42.80}{3.68} & \meanstd{0.41}{0.04} & \meanstd{10.78}{0.51} \\
L2P \cite{wang2022learning} & \meanstd{72.67}{0.95} & \meanstd{59.95}{0.61} & \meanstd{51.28}{1.47} & \meanstd{61.30}{0.46} & \meanstd{37.40}{1.06} & \meanstd{37.80}{0.93} & \meanstd{0.36}{0.02} & \meanstd{3.16}{0.26} \\
DualPrompt \cite{wang2022dualprompt} & \meanstd{64.21}{3.99} & \meanstd{62.42}{0.94} & \underline{\meanstd{55.13}{2.42}} & \meanstd{60.59}{1.99} & \meanstd{41.04}{3.18} & \meanstd{42.05}{2.28} & \meanstd{0.37}{0.03} & \meanstd{7.26}{2.18} \\
S-Prompts \cite{wang2022s} & \underline{\meanstd{78.33}{1.59}} & \meanstd{62.04}{0.55} & \meanstd{41.02}{7.22} & \meanstd{60.47}{1.91} & \meanstd{37.27}{2.05} & \meanstd{38.53}{1.90} & \meanstd{0.38}{0.03} & \meanstd{0.54}{0.15} \\
CODA-Prompt \cite{smith2023coda} & \meanstd{64.12}{9.78} & \meanstd{63.25}{0.61} & \meanstd{53.53}{7.22} & \meanstd{60.30}{1.15} & \meanstd{42.26}{2.11} & \meanstd{42.38}{1.70} & \meanstd{0.38}{0.01} & \meanstd{8.37}{5.89} \\
HiDe-Prompt \cite{wang2025hide} & \meanstd{75.30}{0.38} & \meanstd{63.62}{2.28} & \meanstd{50.00}{5.35} & \underline{\meanstd{62.98}{2.64}} & \underline{\meanstd{55.09}{3.22}} & \underline{\meanstd{53.67}{3.44}} & \underline{\meanstd{0.48}{0.04}} & \meanstd{3.86}{0.34} \\
HiDe-LoRA \cite{wang2025hide} & \meanstd{75.42}{1.75} & \meanstd{61.28}{2.00} & \meanstd{46.15}{2.54} & \meanstd{60.95}{0.91} & \meanstd{53.69}{1.14} & \meanstd{51.90}{1.99} & \meanstd{0.45}{0.00} & \meanstd{4.44}{0.88} \\
DCE \cite{li2025addressing} & \meanstd{74.15}{4.55} & \meanstd{59.57}{3.09} & \meanstd{46.60}{7.33} & \meanstd{60.11}{4.55} & \meanstd{50.40}{4.56} & \meanstd{52.73}{4.40} & \meanstd{0.43}{0.06} & \underline{\meanstd{0.19}{0.16}} \\
\midrule
\textbf{ToRe (Ours)} & \textbf{\meanstd{82.21}{0.47}} & \textbf{\meanstd{68.31}{0.77}} & \textbf{\meanstd{56.41}{4.44}} & \textbf{\meanstd{68.98}{1.68}} & \textbf{\meanstd{56.43}{1.68}} & \textbf{\meanstd{55.25}{1.93}} & \textbf{\meanstd{0.52}{0.02}} & \textbf{\meanstd{0.10}{0.16}} \\
\midrule
Joint Training & \meanstd{83.12}{0.34} & \meanstd{71.42}{0.29} & \meanstd{57.69}{0.96} & \meanstd{70.74}{0.50} & \meanstd{57.55}{0.32} & \meanstd{56.53}{0.82} & \meanstd{0.55}{0.01} & -- \\
\bottomrule
\end{tabular*}
\end{table*}

\section{Experiments}
\label{sec:experiments}

In this section, we evaluate the efficacy of the proposed framework, ToRe. We first detail the experimental setup, including the benchmarks, baselines, evaluation metrics, and implementation protocols in Sec.~\ref{sec:setup}. We then present the main quantitative comparisons against state-of-the-art DIL methods in Sec.~\ref{sec:results}, followed by ablation studies validating our architectural choices and routing reliability on DR in Sec.~\ref{sec:ablation}. Finally, we analyze the gating module from complementary quantitative and qualitative perspectives in Sec.~\ref{sec:gating_module_analysis}.

\subsection{Experimental Setup}
\label{sec:setup}

\subsubsection{Datasets and Benchmarks}
We evaluate ToRe on three domain incremental learning benchmarks constructed from heterogeneous ophthalmic tasks.
The first is {diabetic retinopathy grading}, a 5-class classification task using color fundus photographs (CFP) (Normal, Mild, Moderate, Severe, Proliferative DR) comprising a sequence of three datasets: APTOS~\cite{aptos2019} (India), Messidor-2~\cite{decenciere2014feedback,abramoff2013automated} (France), and IDRiD~\cite{porwal2018indian} (India).
The second is {age-related macular degeneration grading}, a binary classification (Normal \textit{vs.} AMD) task using CFP comprising ODIR~\cite{li2020benchmark} (China), ADAM~\cite{dt4f-rt59-20} (China), and HYAMD~\cite{meisel2025hyamd} (Israel).
The third is an OCT multi-label biomarker prediction task comprising RETOUCH~\cite{bogunovic2019retouch}, AMD-SD~\cite{hu2024amd}, and APTOS-2021~\cite{zhang2026predicting}. The task predicts the presence of intraretinal fluid, subretinal fluid, and pigment epithelial detachment in individual B-scans.
These datasets originate from diverse clinical centers with varied imaging protocols and patient demographics, resulting in significant style variations across domains.
Furthermore, Table~\ref{tab:dataset_stats} details the class and label distributions, highlighting the severe class imbalance and label distribution shift.

\subsubsection{Baselines}
We compare ToRe against five distinct categories of methods:\footnote{For the multi-label OCT benchmark, HiDe-Prompt, HiDe-LoRA, and DCE are not included because their available implementations are designed for single-label classification and would require method-level redesign for this protocol.}
(i) {Standard baselines:} Joint training and sequential full fine-tuning (FT-Seq).
FT-Seq sequentially fine-tunes the full model using the current-domain training data at each stage and serves as a full-parameter sequential reference. Joint Training uses pooled training data from all domains simultaneously and serves as an upper-bound reference for the DIL setting.
(ii) {Classic DIL:} EWC~\cite{kirkpatrick2017overcoming} and LwF~\cite{li2017learning}.
(iii) {Parameter-sharing DIL:} L2P~\cite{wang2022learning}, DualPrompt~\cite{wang2022dualprompt}, CODA-Prompt~\cite{smith2023coda}, and HiDe-PET~\cite{wang2025hide} (including both HiDe-Prompt and HiDe-LoRA variants).
(iv) Parameter-isolation DIL: S-Prompts~\cite{wang2022s}.
(v) Classifier-calibration DIL: DCE~\cite{li2025addressing}.

\subsubsection{Evaluation Metrics}
We utilize seven metrics to assess performance:
(i) {Average accuracy (Avg.ACC):} Measures overall performance on the final model trained on domain $T$, defined as $\frac{1}{T} \sum_{i=1}^{T} A_{T,i}$.
(ii) {Average F1 score (Avg. F1):} Assesses robustness to class imbalance, computed as the average of per-domain Macro-F1 scores: $\frac{1}{T} \sum_{i=1}^{T} \text{F1}_{T,i}$.
(iii) Average Balanced Accuracy (Avg. BAcc): Measures class-balanced performance as the domain average of the mean recall across classes.
(iv) Average Matthews Correlation Coefficient (Avg. MCC): Measures the correlation between predicted and true labels, averaged across domains on its coefficient scale of $[-1,1]$.
(v) Average Macro Average Precision (Avg. Macro AP): Computes Macro Average Precision for each OCT domain and then averages it across domains.
(vi) Average Area Under the Receiver Operating Characteristic Curve (Avg. AUROC): Averages the per-domain AUROC for threshold-independent evaluation on OCT.
(vii) Forgetting: Measures the average reduction in the designated benchmark metric on previous domains after learning the complete sequence. The designated metric is accuracy for DR and AMD. Macro AP is used for OCT, allowing the same stability definition to apply to both single-label and multi-label benchmarks.

\subsubsection{Implementation Details}
All experiments were conducted using the pre-trained RETFound~\cite{zhou2023foundation} backbones (RETFound-CFP for the DR and AMD benchmarks and RETFound-OCT for the OCT benchmark) on one NVIDIA RTX 4090 GPU. Images were resized to $224 \times 224$. We used cross-entropy loss for the single-label DR and AMD benchmarks and binary cross-entropy with logits for OCT, with each OCT target encoded as a multi-hot vector. We employed the AdamW optimizer for ToRe and PEFT baselines, and SGD for EWC and LwF. Training involved a batch size of 64 for 100 epochs with early stopping (patience=10).
For the DR benchmark, we adopted the official data splits provided by RETFound~\cite{zhou2023foundation}. For AMD, we retained the existing held-out test sets and split the remaining training data into training and validation sets using a stratified 90:10 split.\footnote{The ODIR subset comprised normal and AMD-only images, with multilabel cases excluded. All HYAMD images whose filenames matched entries in the annotation CSV were retained.} For OCT, the group-disjoint training, validation, and test splits contain 49, 6, and 15 volumes for RETOUCH; 109, 16, and 31 eyes for AMD-SD; and 154, 22, and 44 patients for APTOS-2021.\footnote{For APTOS-2021, the labeled training release was used after excluding one patient group with conflicting annotations (14 B-scans); the source validation split was excluded because ground-truth annotations were unavailable.} Table~\ref{tab:sample_partitions} gives the image and B-scan counts for these splits. We selected models using the validation sets and evaluated them on the held-out test sets. All experiments were repeated with three random seeds.
\begin{table}[pos=t]
\centering
\caption{Numbers of samples in the training, validation, and test partitions. Counts denote fundus images for DR and AMD and individual B-scans for OCT.}
\label{tab:sample_partitions}
\begingroup
\setlength{\tabcolsep}{2pt}
\footnotesize
\begin{tabular*}{\columnwidth}{@{\extracolsep{\fill}}llrrrr@{}}
\toprule
Benchmark & Dataset & Total used & Training & Validation & Testing \\
\midrule
\multirow{3}{*}{DR} & APTOS 2019 & 3,662 & 2,048 & 514 & 1,100 \\
 & Messidor-2 & 1,744 & 972 & 246 & 526 \\
 & IDRiD & 516 & 329 & 84 & 103 \\
\midrule
\multirow{3}{*}{AMD} & ODIR & 2,497 & 1,798 & 199 & 500 \\
 & ADAM & 800 & 361 & 39 & 400 \\
 & HYAMD & 1,214 & 866 & 95 & 253 \\
\midrule
\multirow{3}{*}{OCT} & RETOUCH & 6,936 & 4,865 & 610 & 1,461 \\
 & AMD-SD & 3,049 & 2,060 & 288 & 701 \\
 & APTOS-2021 & 2,850 & 2,030 & 279 & 541 \\
\bottomrule
\end{tabular*}
\endgroup
\end{table}

For ToRe, LoRA modules (rank=8) were injected into the final 8 blocks with a default recursion depth of $K_{\max}=3$.
To ensure a rigorous comparison, we optimized the hyperparameters for all baselines via grid search on our ophthalmic benchmarks.

\begin{table*}[pos=t]
\centering
\caption{\tablecaption{Performance comparison on the \textbf{AMD Grading} benchmark. Per-domain accuracy and overall metrics are reported as mean (std) over three seeds. Best results are in \textbf{bold}, second best are \underline{underlined}.}}
\label{tab:main_results_amd}
\centering
\shortcites{kirkpatrick2017overcoming,li2017learning,wang2022learning,wang2022dualprompt,wang2022s,smith2023coda,wang2025hide,li2025addressing}
\setlength{\tabcolsep}{0.4pt}
\fontsize{7.5}{9}\selectfont
\let\meanstd\tablestd
\begin{tabular*}{\textwidth}{@{\extracolsep{\fill}}l ccc ccccc@{}}
\toprule
 & \multicolumn{3}{c}{Per-Domain Accuracy (\%)} & \multicolumn{5}{c}{Overall Metrics} \\
\cmidrule(lr){2-4} \cmidrule(lr){5-9}
\tablehead{{Method}} & \tablehead{{ODIR}} & \tablehead{{ADAM}} & \tablehead{{HYAMD}} & \tablehead{{Avg.ACC (\%) $\uparrow$}} & \tablehead{{Avg. F1 (\%) $\uparrow$}} & \tablehead{{Avg. BAcc (\%) $\uparrow$}} & \tablehead{{Avg. MCC $\uparrow$}} & \tablehead{{Forgetting (\%) $\downarrow$}} \\
\midrule
FT-Seq & \meanstd{86.87}{4.44} & \meanstd{81.67}{3.45} & \meanstd{67.06}{2.97} & \meanstd{78.53}{0.65} & \meanstd{61.35}{15.44} & \meanstd{67.12}{14.88} & \meanstd{0.29}{0.25} & \meanstd{8.57}{0.62} \\
EWC \cite{kirkpatrick2017overcoming} & \meanstd{90.13}{4.82} & \meanstd{84.67}{3.21} & \meanstd{65.88}{5.16} & \meanstd{80.23}{2.67} & \meanstd{62.62}{7.15} & \meanstd{61.82}{5.46} & \meanstd{0.34}{0.08} & \meanstd{4.47}{2.97} \\
LwF \cite{li2017learning} & \underline{\meanstd{92.87}{1.33}} & \meanstd{84.00}{4.39} & \meanstd{66.80}{1.04} & \meanstd{81.22}{1.73} & \meanstd{66.23}{11.86} & \meanstd{65.01}{9.08} & \meanstd{0.41}{0.15} & \meanstd{5.38}{2.63} \\
L2P \cite{wang2022learning} & \meanstd{90.60}{0.20} & \meanstd{82.42}{3.51} & \meanstd{66.93}{0.82} & \meanstd{79.98}{1.35} & \meanstd{55.72}{4.72} & \meanstd{56.23}{3.06} & \meanstd{0.21}{0.07} & \meanstd{1.46}{1.44} \\
DualPrompt \cite{wang2022dualprompt} & \meanstd{91.53}{0.12} & \meanstd{81.08}{3.61} & \meanstd{64.95}{0.60} & \meanstd{79.19}{1.42} & \meanstd{51.89}{5.98} & \meanstd{54.11}{3.21} & \meanstd{0.15}{0.07} & \meanstd{1.37}{1.08} \\
S-Prompts \cite{wang2022s} & \meanstd{91.27}{0.58} & \meanstd{78.50}{1.30} & \meanstd{57.31}{1.04} & \meanstd{75.69}{0.30} & \meanstd{49.25}{2.65} & \meanstd{51.78}{2.08} & \meanstd{0.05}{0.08} & \underline{\meanstd{0.17}{0.13}} \\
CODA-Prompt \cite{smith2023coda} & \meanstd{90.53}{1.33} & \meanstd{80.67}{3.11} & \meanstd{66.27}{0.91} & \meanstd{79.16}{0.79} & \meanstd{54.59}{5.29} & \meanstd{55.41}{3.40} & \meanstd{0.18}{0.08} & \meanstd{1.90}{1.24} \\
HiDe-Prompt \cite{wang2025hide} & \meanstd{91.47}{1.30} & \meanstd{84.58}{0.80} & \meanstd{64.30}{1.50} & \meanstd{80.12}{0.35} & \meanstd{73.48}{0.71} & \meanstd{78.29}{0.56} & \meanstd{0.49}{0.01} & \meanstd{3.93}{0.38} \\
HiDe-LoRA \cite{wang2025hide} & \meanstd{89.67}{1.70} & \underline{\meanstd{87.33}{1.01}} & \underline{\meanstd{69.70}{1.99}} & \underline{\meanstd{82.23}{0.96}} & \underline{\meanstd{74.56}{0.82}} & \underline{\meanstd{78.56}{0.13}} & \underline{\meanstd{0.51}{0.01}} & \meanstd{3.23}{1.11} \\
DCE \cite{li2025addressing} & \meanstd{81.40}{1.71} & \meanstd{80.42}{4.47} & \meanstd{60.21}{0.99} & \meanstd{74.01}{0.65} & \meanstd{56.85}{5.37} & \meanstd{57.52}{5.58} & \meanstd{0.16}{0.10} & \meanstd{0.37}{0.14} \\
\midrule
\textbf{ToRe (Ours)} & \textbf{\meanstd{96.00}{0.53}} & \textbf{\meanstd{91.58}{0.63}} & \textbf{\meanstd{70.09}{1.65}} & \textbf{\meanstd{85.89}{0.79}} & \textbf{\meanstd{80.27}{1.00}} & \textbf{\meanstd{79.30}{0.85}} & \textbf{\meanstd{0.61}{0.02}} & \textbf{\meanstd{0.03}{0.06}} \\
\midrule
Joint Training & \meanstd{97.00}{0.00} & \meanstd{89.67}{0.29} & \meanstd{69.87}{1.05} & \meanstd{85.51}{0.25} & \meanstd{80.11}{0.11} & \meanstd{79.48}{0.06} & \meanstd{0.61}{0.00} & -- \\
\bottomrule
\end{tabular*}
\end{table*}

\subsection{Main Results}
\label{sec:results}

\subsubsection{Overall Performance}
\begin{figure}[pos=t!]
\centering
\includegraphics[width=\columnwidth]{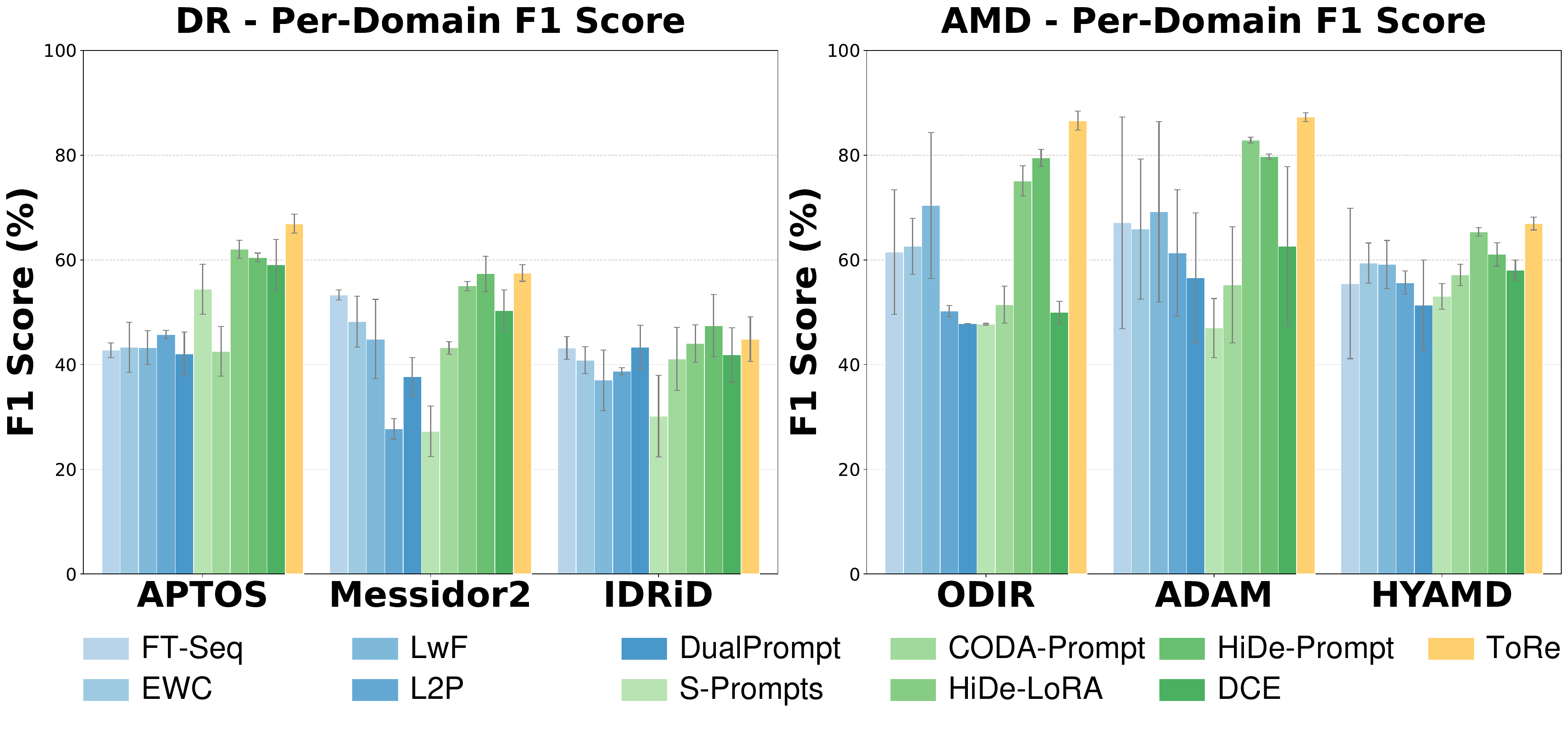}
\caption{Per-domain Macro-F1 scores on the DR and AMD benchmarks, evaluated after the final task. \textbf{(Left)} Comparison on the DR domains. \textbf{(Right)} Comparison on the AMD domains.}
\label{fig:per_domain_f1}
\end{figure}

Table~\ref{tab:main_results_dr} and Table~\ref{tab:main_results_amd} summarize the quantitative comparison on the DR and AMD benchmarks, respectively. Table~\ref{tab:r3_7_oct_extension} presents the OCT multi-label biomarker prediction results. Fig.~\ref{fig:per_domain_f1} further compares per-domain Macro-F1 scores on DR and AMD after learning the complete domain sequence. Joint Training is included as an upper-bound reference for the DIL setting.

\begin{table*}[pos=t]
\begingroup
\centering
\caption{\tablecaption{Performance comparison on the OCT multi-label biomarker prediction benchmark. Per-domain Macro AP and overall metrics are reported as mean (std) over three seeds. Best results are in \textbf{bold}, second best are \underline{underlined}.}}
\label{tab:r3_7_oct_extension}
\centering
\shortcites{kirkpatrick2017overcoming,li2017learning,wang2022learning,wang2022dualprompt,smith2023coda,wang2022s}
\setlength{\tabcolsep}{2pt}
\makebox[\textwidth][c]{%
\footnotesize
\let\meanstd\tablestd
\begin{tabular*}{\textwidth}{@{\extracolsep{\fill}}l ccc ccc@{}}
\toprule
 & \multicolumn{3}{c}{Per-Domain Macro AP (\%) $\uparrow$} & \multicolumn{3}{c}{Overall Metrics} \\
\cmidrule(lr){2-4} \cmidrule(lr){5-7}
\tablehead{{Method}} & \tablehead{{RETOUCH}} & \tablehead{{AMD-SD}} & \tablehead{{APTOS-2021}} & \tablehead{{Avg. Macro AP (\%) $\uparrow$}} & \tablehead{{Avg. AUROC (\%) $\uparrow$}} & \tablehead{{Forgetting (\%) $\downarrow$}} \\
\midrule
FT-Seq & \meanstd{89.26}{0.03} & \meanstd{74.25}{2.44} & \underline{\meanstd{86.36}{0.29}} & \meanstd{83.29}{0.73} & \meanstd{86.86}{0.66} & \meanstd{12.74}{1.16} \\
EWC \cite{kirkpatrick2017overcoming} & \meanstd{90.13}{0.23} & \meanstd{73.94}{0.93} & \meanstd{84.73}{0.40} & \meanstd{82.93}{0.23} & \meanstd{85.06}{0.58} & \meanstd{11.24}{0.55} \\
LwF \cite{li2017learning} & \underline{\meanstd{91.02}{0.22}} & \meanstd{80.38}{0.83} & \meanstd{85.29}{0.65} & \meanstd{85.56}{0.55} & \meanstd{87.26}{0.46} & \meanstd{6.76}{0.33} \\
L2P \cite{wang2022learning} & \meanstd{50.60}{1.55} & \meanstd{78.68}{0.68} & \meanstd{54.14}{2.94} & \meanstd{61.14}{1.34} & \meanstd{70.80}{0.50} & \meanstd{1.36}{0.53} \\
DualPrompt \cite{wang2022dualprompt} & \meanstd{86.04}{1.12} & \meanstd{74.42}{4.90} & \meanstd{83.24}{2.53} & \meanstd{81.23}{1.26} & \meanstd{86.18}{0.54} & \meanstd{7.26}{2.36} \\
CODA-Prompt \cite{smith2023coda} & \meanstd{88.44}{1.48} & \meanstd{72.65}{3.20} & \meanstd{82.64}{0.73} & \meanstd{81.24}{0.60} & \meanstd{86.54}{0.49} & \meanstd{7.51}{1.24} \\
S-Prompts \cite{wang2022s} & \meanstd{90.27}{1.84} & \underline{\meanstd{87.55}{0.13}} & \meanstd{80.31}{1.18} & \underline{\meanstd{86.04}{0.50}} & \underline{\meanstd{89.58}{0.28}} & \underline{\meanstd{0.31}{0.30}} \\
\midrule
\textbf{ToRe (Ours)} & \textbf{\meanstd{94.48}{0.43}} & \textbf{\meanstd{91.79}{0.47}} & \textbf{\meanstd{90.46}{0.47}} & \textbf{\meanstd{92.24}{0.08}} & \textbf{\meanstd{93.53}{0.07}} & \textbf{\meanstd{0.21}{0.08}} \\
\midrule
Joint Training & \meanstd{96.22}{0.51} & \meanstd{92.95}{0.48} & \meanstd{91.05}{0.36} & \meanstd{93.40}{0.19} & \meanstd{94.69}{0.13} & -- \\
\bottomrule
\end{tabular*}%
}
\endgroup
\end{table*}

ToRe improves Avg. ACC by 6.00 percentage points over HiDe-Prompt on DR ($p = 0.0204$) and by 3.66 percentage points over HiDe-LoRA on AMD ($p = 0.0178$). On OCT, ToRe improves Avg. Macro AP and Avg. AUROC by 6.20 and 3.95 percentage points, respectively, over S-Prompts, with $p = 0.00219$ for Avg. Macro AP. These p-values are from exploratory two-sided paired t-tests over three matched seeds.

ToRe also performs consistently well under class imbalance, achieving the highest mean Avg. F1, Avg. BAcc, and Avg. MCC among the compared DIL methods on both DR and AMD. The Avg. F1 gain is particularly marked on AMD, reaching 5.71 percentage points over HiDe-LoRA.

Standard fine-tuning (FT-Seq) suffers from severe forgetting (AF: 13.25\% on DR), confirming the inherent challenge of DIL. PEFT baselines generally reduce this effect, whereas ToRe maintains low forgetting across all three benchmarks, with 0.10\% on DR, 0.03\% on AMD, and 0.21\% on OCT.

\color{black}

\subsection{Ablation Studies}
\label{sec:ablation}

\subsubsection{Component Analysis}

Table~\ref{tab:r2_2_3b_mechanism} presents a component-wise study of token-adaptive recursion on the DR benchmark. We compare uniform computation without gating, learned token selection, and iteration embedding (IE).

The two no-gate settings show that uniformly increasing recursion from one to three steps does not improve Avg. F1 or Avg. BAcc. By contrast, learned token selection at three steps improves Avg. F1 by 5.29 percentage points and Avg. BAcc by 4.75 percentage points over the no-gate setting. This contrast supports learning which tokens require deeper recursive processing. Adding IE further improves these mean scores by 2.88 and 2.65 percentage points, respectively, giving Full ToRe the highest mean scores across all four metrics. This additional gain is consistent with IE helping the shared gate distinguish recursion steps.

\begin{table}[pos=t]
\begingroup
\centering
\caption{\tablecaption{Ablation study of token-adaptive recursion on the DR test set. Results are reported as mean (std) over three seeds. IE denotes iteration embedding. Best means are in \textbf{bold}; second-best means are \underline{underlined}.}}
\label{tab:r2_2_3b_mechanism}
\label{tab:ablation_components}
\setlength{\tabcolsep}{0.5pt}
\fontsize{7.5}{9}\selectfont
\renewcommand{\tablehead}[1]{{\fontsize{6}{7}\selectfont #1}}
\let\meanstd\tablestd
\begin{tabular*}{\columnwidth}{@{\extracolsep{\fill}}lcccc@{}}
\toprule
\tablehead{{Setting}} & \tablehead{{Avg. ACC (\%) $\uparrow$}} & \tablehead{{Avg. F1 (\%) $\uparrow$}} & \tablehead{{Avg. BAcc (\%) $\uparrow$}} & \tablehead{{Avg. MCC $\uparrow$}} \\
\midrule
w/o gate, 1 step & \meanstd{65.34}{1.27} & \meanstd{49.13}{2.43} & \meanstd{48.49}{2.43} & \meanstd{0.47}{0.02} \\
w/o gate, 3 steps & \meanstd{66.61}{2.16} & \meanstd{48.26}{2.09} & \meanstd{47.85}{1.58} & \meanstd{0.49}{0.03} \\
w/ gate, 3 steps & \underline{\meanstd{67.55}{1.63}} & \underline{\meanstd{53.55}{2.40}} & \underline{\meanstd{52.60}{2.30}} & \underline{\meanstd{0.50}{0.02}} \\
Full ToRe (+IE) & \textbf{\meanstd{68.98}{1.68}} & \textbf{\meanstd{56.43}{1.68}} & \textbf{\meanstd{55.25}{1.93}} & \textbf{\meanstd{0.52}{0.02}} \\
\bottomrule
\end{tabular*}
\endgroup
\end{table}

\subsubsection{Sensitivity to Recursion Depth ($K_{\max}$)}

We next examine sensitivity to maximum recursion depth on the DR test set. Table~\ref{tab:ablation_kmax} reports the results for $K_{\max}=1,3,5,$ and $10$. Our default $K_{\max}=3$ achieves the highest mean scores across all four metrics. Across these settings, ToRe maintains strong predictive performance, with Avg. F1 ranging from 52.48\% to 56.43\% and Avg. BAcc from 51.74\% to 55.25\%.

\begin{table}[pos=t]
\begingroup
\centering
\caption{\tablecaption{Sensitivity to maximum recursion depth $K_{\max}$ on the DR test set. Results are reported as mean (std) over three seeds. Best means are in \textbf{bold}; second-best means are \underline{underlined}.}}
\label{tab:ablation_kmax}
\setlength{\tabcolsep}{1pt}
\footnotesize
\let\meanstd\tablestd
\begin{tabular*}{\columnwidth}{@{\extracolsep{\fill}}lcccc@{}}
\toprule
\tablehead{{$K_{\max}$}} & \tablehead{{Avg. ACC (\%) $\uparrow$}} & \tablehead{{Avg. F1 (\%) $\uparrow$}} & \tablehead{{Avg. BAcc (\%) $\uparrow$}} & \tablehead{{Avg. MCC $\uparrow$}} \\
\midrule
1 & \meanstd{66.77}{0.67} & \meanstd{52.48}{0.50} & \meanstd{51.74}{0.86} & \meanstd{0.49}{0.01} \\
3 & \textbf{\meanstd{68.98}{1.68}} & \textbf{\meanstd{56.43}{1.68}} & \textbf{\meanstd{55.25}{1.93}} & \textbf{\meanstd{0.52}{0.02}} \\
5 & \underline{\meanstd{67.98}{2.50}} & \underline{\meanstd{53.82}{3.65}} & \underline{\meanstd{53.88}{3.78}} & \underline{\meanstd{0.51}{0.03}} \\
10 & \meanstd{67.66}{0.60} & \meanstd{52.91}{0.88} & \meanstd{52.79}{0.98} & \meanstd{0.50}{0.01} \\
\bottomrule
\end{tabular*}
\endgroup
\end{table}

\subsubsection{\texorpdfstring{Sensitivity to LoRA Rank and Insertion Depth}{Sensitivity to LoRA Rank and Insertion Depth}}

To assess whether the default ToRe configuration is sensitive to adaptation capacity and insertion depth, we evaluate LoRA rank and the number of recursive blocks on the DR benchmark using Avg. F1 and Avg. BAcc.

\begin{figure}[pos=t]
\begingroup
\centering
\includegraphics[width=\columnwidth]{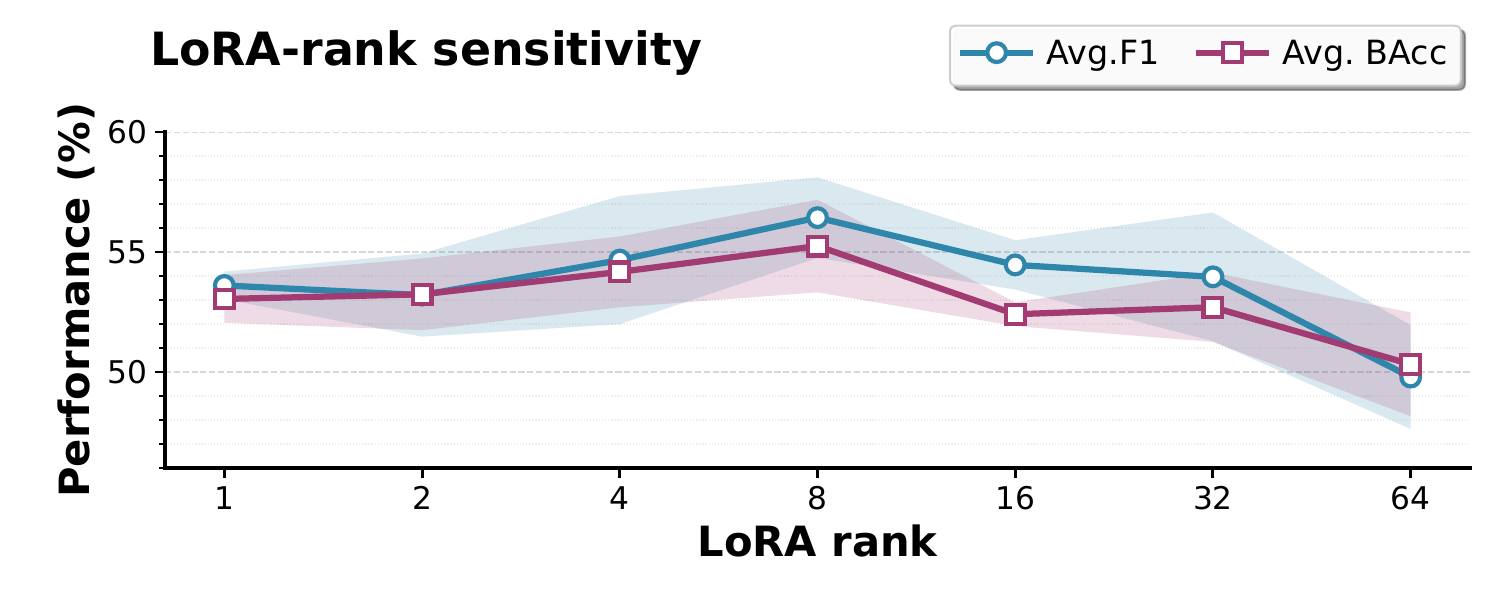}
\caption{Sensitivity of Avg. F1 and Avg. BAcc to LoRA rank on the DR test set over three seeds. Shaded bands show one standard deviation.}
\label{fig:lora_rank_sensitivity}
\endgroup
\end{figure}

Avg. F1 and Avg. BAcc across different LoRA ranks are reported in Fig.~\ref{fig:lora_rank_sensitivity}. Rank 8, our default setting, achieves the highest mean scores on both metrics among the tested settings, whereas rank 64 yields the lowest.

\begin{figure}[pos=t]
\begingroup
\centering
\includegraphics[width=\columnwidth]{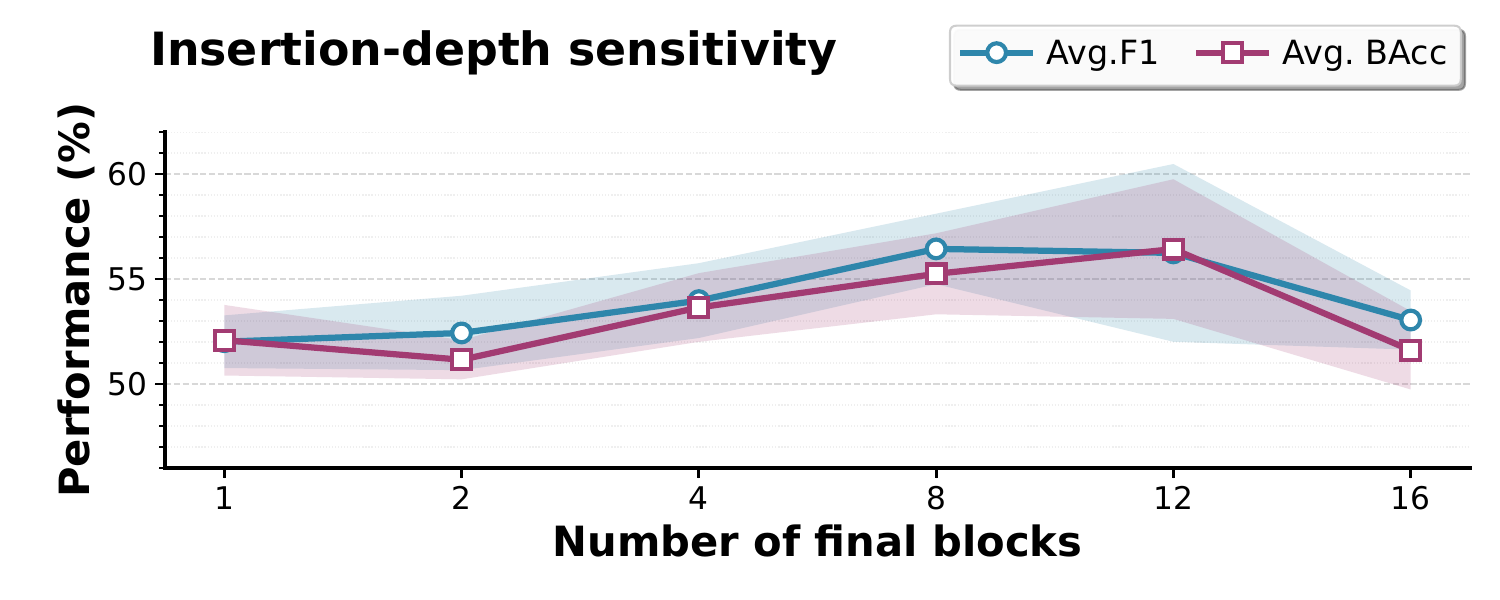}
\caption{Sensitivity of Avg. F1 and Avg. BAcc to insertion depth on the DR test set over three seeds. Shaded bands show one standard deviation.}
\label{fig:insertion_depth_sensitivity}
\endgroup
\end{figure}

Avg. F1 and Avg. BAcc across different insertion depths are reported in Fig.~\ref{fig:insertion_depth_sensitivity}. Among the settings shown, using the final eight blocks achieves the highest Avg. F1, whereas using twelve blocks achieves the highest Avg. BAcc. Both metrics decrease when the insertion depth increases to sixteen blocks. We retain eight recursive blocks as the default configuration, maintaining strong predictive performance while limiting the number of blocks involved in recursion.

\subsubsection{\texorpdfstring{Routing Reliability}{Routing Reliability}}
\label{sec:routing_reliability}

We further evaluate domain identification on the DR, AMD, and OCT benchmarks by measuring routing accuracy and examining the diagnostic impact of route mismatches.

\begin{figure}[pos=t]
\begingroup
\centering
\includegraphics[width=\columnwidth]{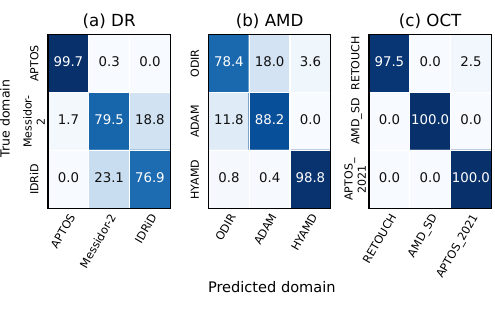}
\caption{Row-normalized domain-routing confusion matrices (\%) on the (a) DR, (b) AMD, and (c) OCT test sets. Rows indicate true domains, and columns indicate predicted domains, in the same domain order within each panel.}
\label{fig:routing_reliability_dr}
\endgroup
\end{figure}

\begin{figure*}[pos=t]
\begingroup
\centering
\includegraphics[width=\textwidth]{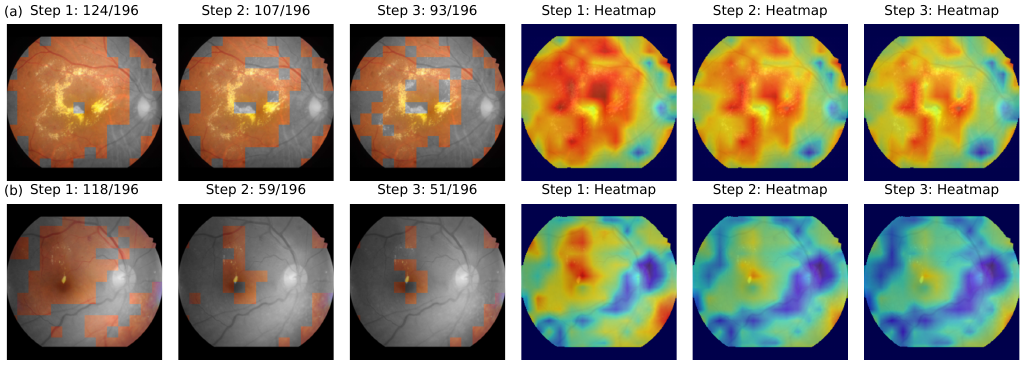}
\caption{\textbf{Evolution of active-token selection across recursive steps.} Rows (a) and (b) show two examples. Columns 1--3 show active-token overlays at Steps 1--3, with the number of selected tokens above each image; Columns 4--6 show the corresponding gating activation maps.}
\label{fig:router_viz}
\endgroup
\end{figure*}

\textit{\textbf{Domain Identification Accuracy:}} Overall Route Accuracy is the proportion of all test images assigned to the correct domain. Macro Route Accuracy first computes this accuracy separately for each domain and then averages the domain-specific values equally. Overall and Macro Route Accuracy are 92.20\% and 85.37\% on DR, 86.30\% and 88.49\% on AMD, and 98.67\% and 99.18\% on OCT, respectively. Fig.~\ref{fig:routing_reliability_dr} shows the row-normalized confusion matrices. Most DR errors occur between Messidor-2 and IDRiD, while AMD errors mainly involve ODIR and ADAM.

\textit{\textbf{Impact of Route Mismatch:}} To determine how domain mismatches affect diagnostic performance, we evaluate all misrouted images under both the predicted route and their true-domain route. The predicted route uses the module selected by domain identification, whereas the true-domain route uses the module associated with the image's true domain. The true-domain route is used only as an analysis reference because it is unavailable during inference.

Table~\ref{tab:routing_diagnostic_impact} shows that ToRe achieves slightly lower mean ACC under predicted routing than under true-domain routing on the misrouted DR and AMD images, with gaps below 0.5 percentage points.

\begingroup
\emergencystretch=1em
On OCT, all 36 route mismatches occurred from RETOUCH to APTOS-2021. The lower accuracy under predicted routing mainly results from additional false positives on these images. In particular, predicted routing produces 7 to 15 IRF false positives per seed, compared with none under true-domain routing. Overall, predicted routing yields 6 to 14 fewer fully correct predictions per seed, corresponding to a mean decrease of 28.70 percentage points in subset accuracy. This drop highlights the importance of selecting the appropriate domain-specific module for biomarker prediction. In future work, we will explore how to improve domain identification for these mismatch cases.
\par
\endgroup

\begin{table}[pos=t]
\begingroup
\centering
\caption{\tablecaption{Diagnostic performance on misrouted test images under the predicted and true-domain routes. Results are reported as mean (std) over three seeds. Misrouted (\%) indicates the percentage of each benchmark's test set assigned to an incorrect domain. Pred. and True denote the predicted and true-domain routes, respectively. Acc. denotes ACC for DR and AMD and subset accuracy for OCT, which requires all three biomarker labels to be correct. $\Delta$ denotes true-domain minus predicted-route accuracy in percentage points.}}
\label{tab:routing_diagnostic_impact}
\setlength{\tabcolsep}{1pt}
\footnotesize
\let\meanstd\tablestd
\begin{tabular*}{\columnwidth}{@{\extracolsep{\fill}}lcccc@{}}
\toprule
\tablehead{{Benchmark}} & \tablehead{{Misrouted (\%)}} & \tablehead{{Pred. Acc. (\%)}} & \tablehead{{True Acc. (\%)}} & \tablehead{{$\Delta$ (pp)}} \\
\midrule
DR & 7.80 & \meanstd{63.21}{1.13} & \meanstd{63.70}{1.48} & \meanstd{0.49}{2.60} \\
AMD & 13.70 & \meanstd{89.87}{1.27} & \meanstd{90.30}{0.73} & \meanstd{0.42}{0.73} \\
OCT & 1.33 & \meanstd{56.48}{11.23} & \meanstd{85.19}{3.21} & \meanstd{28.70}{11.23} \\
\bottomrule
\end{tabular*}
\endgroup
\end{table}

\subsection{\texorpdfstring{Gating Module Analysis}{Gating Module Analysis}}
\label{sec:gating_module_analysis}

We analyze the gating module from complementary quantitative and qualitative perspectives. The recursion-depth statistics measure how recursive computation is allocated across tokens, the lesion token analysis evaluates selection behavior against external pathological masks, and the visualization shows how the selected tokens evolve across recursive steps.

\subsubsection{\texorpdfstring{Quantitative Analysis}{Quantitative Analysis}}

Mean Recursion Depth measures the number of recursive updates received by each token at each recursive block. ``All tokens'' averages this count over every token, whereas ``selected tokens'' includes only tokens selected for at least one recursive update. The statistics pool tokens across all test images and recursive blocks within each seed; the class token is excluded. Table~\ref{tab:gating_recursion_depth} reports the mean and standard deviation over three seeds.

\begin{table}[pos=!htbp]
\begingroup
\centering
\caption{\tablecaption{Mean recursion depth on the DR, AMD, and OCT test sets. Results are reported as mean (standard deviation) over three seeds.}}
\label{tab:gating_recursion_depth}
\let\meanstd\tablestd
\begin{tabular*}{\columnwidth}{@{\extracolsep{\fill}}lcc@{}}
\toprule
{Benchmark} & {All tokens} & {Selected tokens} \\
\midrule
DR & \meanstd{1.37}{0.32} & \meanstd{2.70}{0.08} \\
AMD & \meanstd{1.15}{0.08} & \meanstd{2.69}{0.08} \\
OCT & \meanstd{1.27}{0.45} & \meanstd{2.61}{0.21} \\
\bottomrule
\end{tabular*}
\endgroup
\end{table}

Across all three benchmarks, selected tokens receive more than two recursive updates on average, whereas the mean over all tokens is lower. This difference shows that the gate concentrates recursive computation on a selected subset.

To examine the gating module more directly, we use the lesion-segmentation dataset provided by the IDRiD challenge, which contains pixel-level annotations for four lesion types: haemorrhages, hard exudates, microaneurysms, and soft exudates~\cite{porwal2018indian}. For each image, we compare the mean selection rates of lesion and non-lesion tokens across recursive blocks, recursion steps, and three seeds. We also compare the mean selection rates of the two groups at each block-step combination across all images and seeds. The mean selection rate is higher for lesion tokens in 80 of 81 images (98.77\%) and 22 of 24 block-step combinations (91.67\%). Across three seeds, the mean selection rates are 45.30\% for lesion tokens and 38.74\% for non-lesion tokens, with a mean difference of \meanstd{6.56}{2.53} percentage points. These results show that the gating module more frequently selects lesion tokens for recursive processing. This preference is consistent with our motivation to allocate additional computation to complex tokens, such as lesion-associated tokens.

\subsubsection{\texorpdfstring{Qualitative Analysis}{Qualitative Analysis}}

The quantitative results above characterize the overall token-selection behavior, while Fig.~\ref{fig:router_viz} illustrates how the active-token set evolves across recursive steps in individual examples. The upper example, which contains more spatially extensive lesions, retains a broader active set, with 124, 107, and 93 tokens selected at Steps 1--3, respectively. In contrast, the active-token count in the lower example decreases more rapidly from 118 to 59 and 51. Thus, although the active set contracts in both examples as recursion proceeds, the different contraction patterns show that the gating module adaptively narrows recursive computation according to the input.

\color{black}

\section{Discussion}
\label{sec:discussion}

Ophthalmic DIL involves a compound setting in which within-domain class imbalance coexists with across-domain changes in label distributions and imaging styles; its evaluation should therefore consider both imbalance-aware final diagnostic performance and how well previously learned knowledge is preserved.

Effective DIL requires strong final diagnostic performance after learning all domains while keeping forgetting low, because lower forgetting indicates that knowledge learned from earlier domains is better preserved~\cite{lopez2017gradient,chaudhry2019continual}. This joint criterion makes S-Prompts~\cite{wang2022s} a useful reference for ToRe: both methods use domain-specific parameter isolation, which reduces direct cross-domain optimization interference and is consistent with their low forgetting on DR and AMD. However, their class-balanced diagnostic performance differs substantially on the class-imbalanced DR and AMD benchmarks (Table~\ref{tab:dataset_stats}). S-Prompts reaches Avg. BAcc values of 38.53\% and 51.78\%, respectively, whereas ToRe reaches 55.25\% and 79.30\% (Tables~\ref{tab:main_results_dr} and~\ref{tab:main_results_amd}). ToRe also shows lower mean forgetting on DR (0.10\% vs. 0.54\%) and AMD (0.03\% vs. 0.17\%). On OCT, ToRe achieves a higher Avg. Macro AP than S-Prompts (92.24\% vs. 86.04\%), while both methods maintain low forgetting (0.21\% vs. 0.31\%; Table~\ref{tab:r3_7_oct_extension}).

ToRe is designed for imbalanced ophthalmic DIL by coordinating complementary roles within one framework. During current-domain training, parameter isolation confines optimization to the current module while the frozen backbone and earlier modules remain fixed, and token-adaptive recursion refines token representations within that module. At inference, domain identification uses the stored domain keys to retrieve the corresponding module, within which token-adaptive recursion refines token representations. Together, these components enable ToRe to achieve strong class-balanced performance while maintaining low forgetting on the imbalanced ophthalmic DIL benchmarks.

Alongside these diagnostic benefits, ToRe uses a small number of learned parameters and has moderate inference memory requirements. Across the three benchmarks, its accumulated learned parameters average 0.92M, approximately 0.30\% of the full model, while peak allocated GPU memory averages 1.63 GiB. These results complement the diagnostic performance reported in Tables~\ref{tab:main_results_dr}, \ref{tab:main_results_amd}, and~\ref{tab:r3_7_oct_extension}.

Table~\ref{tab:efficiency_profile} summarizes the parameter and computational costs. We evaluate final-stage checkpoints on an NVIDIA RTX 4090 with batch size 32 and AMP FP16. Measurements include domain identification and model forward computation, excluding data loading, host-to-device transfer, and metric computation.

\begin{table}[pos=!t]
\begingroup
\centering
\caption{\tablecaption{Parameter and computational cost comparison. Computational costs are means over three seeds, averaged equally across DR, AMD, and OCT; $\dagger$ indicates averages over DR and AMD only. Params denotes the reported trainable parameters (M), FLOPs is estimated per image (G/image), Throughput is reported in images/s, and Memory denotes peak allocated GPU memory (GiB).}}
\label{tab:efficiency_profile}
\label{tab:efficiency_profile_dr}
\label{tab:efficiency_profile_amd}
\setlength{\tabcolsep}{2pt}
\begin{tabular*}{\columnwidth}{@{\extracolsep{\fill}}lcccc@{}}
\toprule
{Method} & {Params} & {FLOPs} & {Throughput} & {Memory} \\
\midrule
FT-Seq & 303.31 & 119.61 & 131.36 & 1.890 \\
EWC & 303.31 & 119.61 & 103.82 & 2.649 \\
LwF & 303.31 & 119.61 & 125.97 & 1.896 \\
L2P & 0.12 & 279.27 & 51.35 & 3.441 \\
DualPrompt & 0.97 & 247.01 & 52.78 & 3.440 \\
S-Prompts & 0.09 & 259.79 & 49.57 & 1.487 \\
CODA-Prompt & 16.90 & 239.25 & 58.79 & 1.421 \\
HiDe-Prompt$\dagger$ & 2.17 & 247.22 & 46.07 & 4.992 \\
HiDe-LoRA$\dagger$ & 7.49 & 325.39 & 38.32 & 5.351 \\
DCE$\dagger$ & 14.28 & 133.14 & 87.74 & 1.514 \\
ToRe & 0.92 & 354.18 & 20.68 & 1.630 \\
\bottomrule
\end{tabular*}
\endgroup
\end{table}

However, the small number of learned parameters does not remove the computational cost of recursion. Although the same parameters are reused within each recursive block, repeated computation results in an average forward cost of 354.18 GFLOPs per image. Throughput was measured with the FLOPs profiler enabled. ToRe achieves a throughput of 20.68 images/s, reflecting a trade-off between additional recursive computation and inference speed. Future work will investigate more efficient execution of the recursion loop, with the aim of reducing redundant computation and improving throughput while preserving diagnostic performance.

Beyond computational cost, a limitation of this work is that the gating module learns to select tokens via the optimization of the classification loss, lacking explicit guidance on what constitutes a ``lesion signature.'' This implicit learning process may not always align perfectly with ophthalmic semantics. Future work will explore multimodal integration. By incorporating clinical text reports as auxiliary supervision, we aim to explicitly guide the gating module to associate visual tokens with ophthalmic concepts (e.g., ``hemorrhage,'' ``exudates''). This could help the model learn policies that are semantically aligned with ophthalmic knowledge, potentially improving interpretability and robustness to complex disease cases.
Building on this semantic guidance, future work could introduce task-specific spatial supervision and dense-prediction heads to extend the token-adaptive recursion mechanism from image-level diagnosis to lesion localization and segmentation.

A further limitation concerns how ToRe handles new domains in more complex data streams. The current method assumes that the training samples from each incoming domain are provided together. These samples are used to train a new domain-specific module and compute its domain key while the frozen backbone and earlier modules remain fixed; the domain key and its associated module are then added to the module bank. At inference, a test query is compared with the stored domain keys to select a module, so no domain label is required for the test image. However, the method does not automatically determine when a new domain emerges or which samples belong to it in mixed or gradually shifting streams. Future work will explore uncertainty-aware routing, open-set domain identification, and automatic detection of new domains.

\section{Conclusion}
\label{sec:conclusion}
In this work, we present ToRe, a parameter-efficient framework for imbalanced ophthalmic DIL under label distribution shift and style variations across domains.
ToRe coordinates a parameter isolation strategy and domain identification to support domain-specific adaptation, while token-adaptive recursion refines the representations of selected tokens through the recursion loop.
Extensive experiments across three ophthalmic DIL benchmarks demonstrate that ToRe achieves strong diagnostic performance under class imbalance while maintaining low forgetting.
Future work will investigate more flexible domain routing and explore extensions of ToRe to volumetric, multimodal, and dense-prediction ophthalmic settings, including lesion localization and segmentation.

\section*{Declaration of generative AI and AI-assisted technologies in the manuscript preparation process}
During the preparation of this work, the authors used ChatGPT in order to polish the English language. After using this tool, the authors reviewed and edited the content as needed and take full responsibility for the content of the published article.

\bibliographystyle{cas-model2-names}

\bibliography{references}

\end{document}